\documentclass[fleqn,10pt]{wlscirep}
\usepackage[utf8]{inputenc}
\usepackage[T1]{fontenc}

\usepackage{multirow}
\usepackage{colortbl}
\usepackage{subcaption}
\usepackage{hhline}

\title{Human action recognition with a large-scale brain-inspired photonic computer}

\author[1,*]{Piotr Antonik}
\author[1]{Nicolas Marsal}
\author[2]{Daniel Brunner}
\author[1,*]{Damien Rontani}
\affil[1]{LMOPS EA 4423 Laboratory, CentraleSup\'elec \& Universit\'e de Lorraine, F-57070 Metz, France}
\affil[2]{FEMTO-ST Institute/Optics Department, CNRS \& University Bourgogne Franche-Comt\'e, F-25030 Besan\c{c}on, France}
\affil[*]{Corresponding authors: piotr.antonik@centralesupelec.fr, damien.rontani@centralesupelec.fr}

\usepackage[]{color}

\newcommand{\pf}[1]{\texttt{#1}} 
\newcommand{\units}[1]{\; \text{#1}}
\newcommand{\rev}[1]{\textcolor{black}{#1}}

\definecolor{dgreen}{rgb}{0.00,0.80,0.40}
\definecolor{dblue}{rgb}{0.00,0.40,0.80}
\definecolor{green100}{rgb}{0.00,0.60,0.00}
\definecolor{green80}{rgb}{0.40,0.80,0.40}
\definecolor{green0}{rgb}{0.70,0.90,0.70}
\definecolor{blue100}{rgb}{0.00,0.33,1.0}
\definecolor{blue96}{rgb}{0.04,0.36,1.0}
\definecolor{blue84}{rgb}{0.16,0.44,1.0}
\definecolor{blue80}{rgb}{0.20,0.47,1.0}
\definecolor{blue76}{rgb}{0.24,0.49,1.0}
\definecolor{blue20}{rgb}{0.80,0.87,1.0}
\definecolor{blue16}{rgb}{0.84,0.89,1.0}
\definecolor{blue12}{rgb}{0.88,0.92,1.0}
\definecolor{blue8}{rgb}{0.92,0.95,1.0}
\definecolor{blue4}{rgb}{0.96,0.97,1.0}
\definecolor{blue0}{rgb}{1.00,1.00,1.0}
\definecolor{red20}{rgb}{0.50,0.30,0.30}
\definecolor{red16}{rgb}{0.60,0.40,0.40}
\definecolor{red12}{rgb}{0.70,0.50,0.50}
\definecolor{red8}{rgb}{0.80,0.60,0.60}
\definecolor{red4}{rgb}{0.90,0.70,0.70}

\newcommand{\g}[2]{\cellcolor{green#2} #1\%}
\newcommand{\rr}[2]{\cellcolor{red#2} #1\%}
\newcommand{\z}{\cellcolor{green0} \phantom{100\%}}

\begin{abstract}
  The recognition of human actions in video streams is a challenging task in computer vision, with cardinal applications in e.g. brain-computer interface and surveillance. Deep learning has shown remarkable results recently, but can be found hard to use in practice, as its training requires large datasets and special purpose, energy-consuming hardware. In this work, we propose a scalable photonic neuro-inspired architecture based on the reservoir computing paradigm, capable of recognising video-based human actions with state-of-the-art accuracy. Our experimental optical setup comprises off-the-shelf components, and implements a large parallel recurrent neural network that is easy to train and can be scaled up to hundreds of thousands of nodes. This work paves the way towards simply reconfigurable and energy-efficient photonic information processing systems for real-time video processing.
\end{abstract}
\begin{document}

\flushbottom
\maketitle
%
%
\thispagestyle{empty}



\section{Introduction}

In recent years, human action recognition has become one of the most popular research areas in the field of computer vision \cite{wu2017recent}.
The driving force of this research field are the potential applications, which can be found in various areas such as surveillance, control, and analysis \cite{moeslund2001survey}. 
Surveillance is concerned with tracking one or several subjects over time and detecting specific actions. 
A typical example is the surveillance of a parking lot for the prevention of car theft.
Applications concerning system-control make use of the captured motions to provide control functionality in games, virtual environments, or to control remote devices \cite{moeslund2001interacting}.
The detailed automatic analysis of motions could be used in clinical studies of e.g. orthopedic patients, or to help athletes improve their performance\cite{moeslund2001survey}.

The recognition of human activities from video sequences is a challenging task due to numerous problems, such as background clutter, partial occlusion, changes in scale or viewpoint, lighting, and appearance \cite{vrigkas2015review}. 
Deep learning, after being successfully applied to speech recognition, natural language processing and recommendation systems, has been recently introduced in the video-based human action recognition research field \cite{wu2017recent}. 
The numerous advantages of these hierarchical approaches -- raw video inputs, automatically deduced features and recognition of complex actions -- attracted much interest from the community. 
However, these approaches also have several drawbacks, such as the need of (very) large datasets, the non-trivial tuning of the hyperparameters, and time- and energy-consuming training process, which commonly requires dedicated high-end hardware such as graphical processing units (GPU).

In this work, we propose an optical signal processing system for classification of video-based human actions.
The idea of optical computing has been investigated for decades as photons propagate without generating heat or signal degradation due to induction and capacitive effects, and thus promise a high level of parallelism in e.g. optical communications. Neural networks would heavily benefit from parallel signal transmission, which, as shown by the increasing usage of optical interconnects in modern computing systems, is one of the strong suits of photonics.
An optical approach could thus allow one to build high-speed and energy-efficient photonic computing devices.

Our experimental optical system implements a shallow recurrent neural network under the so-called \emph{reservoir computing paradigm}.
Reservoir Computing (RC) is a set of machine learning methods for designing and training artificial neural networks \cite{jaeger2004harnessing, maass2002real}. The idea behind these techniques is to exploit the dynamics of a random recurrent neural network to process time series by only training a linear output layer. The resulting system is significantly easier to train: instead of the entire network, only the readout layer is optimised by solving a system of linear equations \cite{lukosevicius2009reservoir}. 
Furthermore, as less parameters are inferred during training, the network can be trained on significantly smaller datasets without the risk of overfitting.
The performance of the numerous experimental implementations of reservoir computing in electronics \cite{appeltant2011information}, optoelectronics \cite{paquot2012optoelectronic,larger2012photonic,martinenghi2012photonic,larger2017high}, optics \cite{duport2012all,brunner2013parallel,vinckier2015high,akrout2016parallel}, and integrated on chip \cite{vandoorne2014experimental} is comparable to other digital algorithms on a series of benchmark tasks, such as wireless channel equalisation \cite{jaeger2004harnessing}, phoneme recognition \cite{triefenbach2010phoneme} and prediction of future evolution of financial time series \cite{NFC}.
Finally, it was shown that the readout layer of photonic reservoir computers can be implemented optically and trained using a digital micro-mirror device\cite{bueno2018reinforcement}.

In this paper, we present an optoelectronic reservoir computer, inspired by Refs. \cite{bueno2018reinforcement,hagerstrom2012experimental}. 
The system is based on the phase modulation of a spatially extended planar wave by means of a spatial light modulator (SLM). 
Our scheme offers a notable parallelisation potential through simultaneous optical processing of the nodes of the reservoir computer, while the physical resolution of the SLM defines the maximal network size. 
This allows for a significantly increased scalability of the network, which is vital for successfully solving the challenging tasks in computer vision.
The experimental setup can accommodate a reservoir of 16,384 nodes, while the physical limitation of the concept is set to as high as $262,144$ neurons.
The input and the output layers, as well as the recurrence of the network, are realised digitally in this work. 

The system is benchmarked on the popular KTH database \cite{schuldt2004recognizing}, which contains video recordings of 6 different motions (walking, jogging, running, boxing, hand waving, and hand clapping) performed by 25 subjects. 
\rev{In particular, we focus on the first scenario ``s1'', containing outdoor videos shot over uniform background}. 
At the pre-processing stage, the histograms of oriented gradients (HOG) algorithm \cite{dalal2005histograms} (described later) is used to extract spatial and shape information from individual video frames.
The photonic reservoir computer is used to classify the 6 motions given the resulting HOG features. 

\rev{The setup is evaluated both experimentally and in simulations. The numerical model was design to mimic the experiment as accurately as possible. It is based on the same nonlinearity, trained and tested on the same data, and the hyperparameters are optimised in the same way.} 
We investigate the scalability of our approach with network sizes ranging from 1,024 to 16,384 nodes and report classification accuracy as high as 92\% which is comparable to the state-of-the-art rates $90.7\% - 95.6\%$ achieved with far more complex and demanding architectures implemented on noiseless digital processors \cite{wu2017recent}. This work thus shows that a challenging computer vision task can be efficiently solved with a simple photonic system. 
It represents a successfull first step towards a video processing system with electronic pre-processing stage (HOG features) and a fully-optical reservoir computer, that benefits from the intrinsic parallelism of photonics, and thus offers a highly-scalable and, potentially, energy-efficient neural network.

\section{Results}
\label{sec:res}

Before presenting the results of this study, we introduce the video-based human action classification task in the context of reservoir computing, and then present the experimental setup. The theory of reservoir computing can be found in the Methods section.

\subsection{Classification of human action with a reservoir computer}
\label{subsec:kth}

\begin{figure}[t]
  \centering
  \includegraphics[width=1\textwidth]{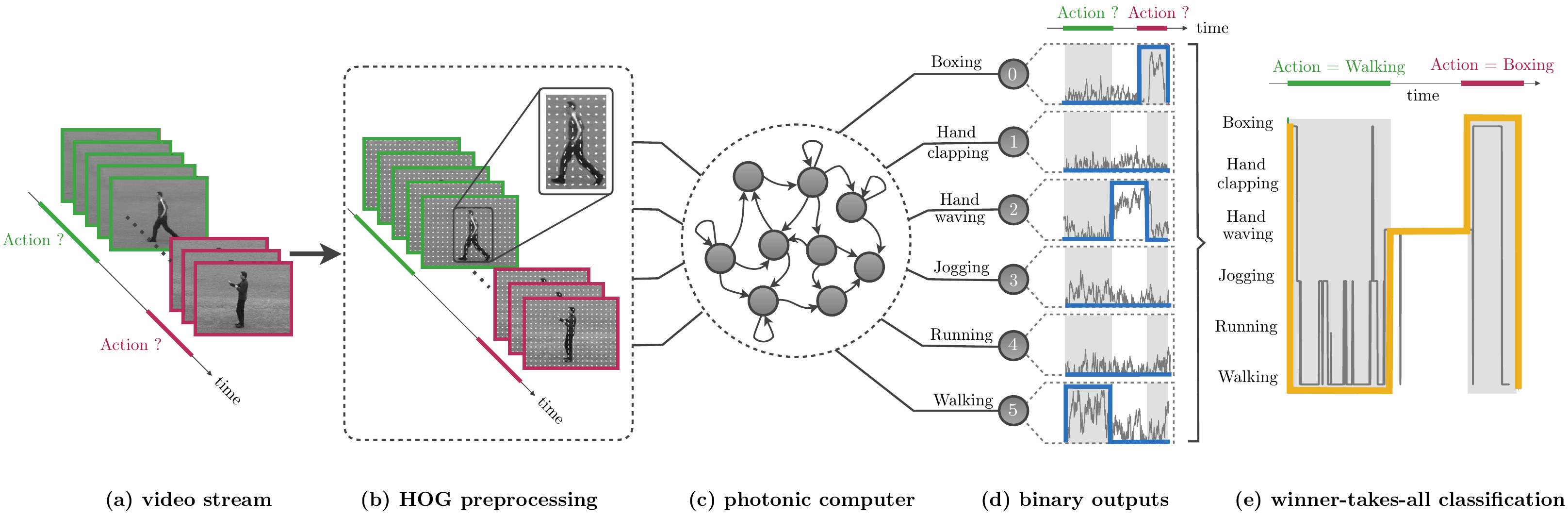}
  \caption{Scheme of principle of how our reservoir computer solves the human action classification task. 
    The video input stream (a) is a concatenation of the 600 video sequences available in the KTH dataset; 450 sequences were used for training, and 150 for testing. 
    The input stream undergoes a pre-processing stage (b), where the HOG algorithm is applied to each individual frame. 
    The dimensionality reduction through the PCA is not illustrated in this figure.
  The selected features are fed into the photonic reservoir computer (c), trained to classify each individual frame. This is achieved by defining 6 binary output nodes (d), one for each action class, that are trained to output 1 for a frame of the corresponding class and 0 for the others. Target outputs are shown in blue. The frame-wise classification (e) is obtained by selecting the node with the maximum output, \textit{i.e.} the winner-takes-all approach. The final decision for a video sequence is given by the class attributed to the most frames of the sequence. \rev{The target class is shown in yellow}. Two examples illustrate the entire process. A boxing sequence, highlighted in red in (a) and (b), is classified unambiguously in (e), as all output nodes in (d) remain low, except for the one corresponding to boxing, that generates a clear spike. A walking sequence, highlighted in green, is more uncertain, as two output nodes -- jogging and walking -- generate high responses in (d). Therefore, the reservoir output (e) oscillates between the two classes \rev{(the faint vertical lines in the light-gray left-hand side region)}. However, since more frames in the sequence are classified as walking ($74.5\%$) than jogging ($23.6\%$), the entire sequence is correctly classified as walking.}
  \label{fig:principle}
\end{figure}

The principles of the human action recognition task in the context of reservoir computing are illustrated in Fig. \ref{fig:principle}.
In this work, we used the popular KTH database of human actions \cite{schuldt2004recognizing}, publicly available online, which consists of video recordings of 6 different motions (walking, jogging, running, boxing, hand waving, and hand clapping) performed by 25 subjects. 
Each subjects performs each motion 4 times, which results in a dataset of 600 video sequences of variables lengths, ranging from 24 to 239 frames.
More details on the video properties of the dataset can be found in the Methods section.
All videos are concatenated together and split into individual frames, giving the raw video stream (see Fig. \ref{fig:principle}(a)), carried forward to the pre-processing stage (Fig. \ref{fig:principle}(b)).

Feature extraction is a common approach in computer vision to provide the classification system, in this case -- a photonic reservoir computer, with the most relevant information. 
We tested our reservoir computer with raw frames, but the classification errors were significantly higher than state of the art. 
Therefore, we turned to the histograms of oriented gradients algorithm, introduced by Dalal and Triggs\cite{dalal2005histograms}.
There, an intra-frame spatial gradient is computed for each pixel and then pooled into one common gradient histogram. Such HOG features
are widely used in computer vision and image processing with the intention of aiding the localisation and detection of objects (see e.g. \cite{bahi2015robust}).
This method is particularly well-suited for pedestrian detection (as well as a variety of common animals and vehicles) in static imagery.
The main idea is that local object appearance and shape can often be expressed well enough by distribution of local intensity gradients or edges' directions.  
The HOG algorithm is further discussed in the Methods section.
To reduce the number of resulting HOG features and simplify computations, we apply the principal component analysis (PCA) \cite{pearson1901liii.,hotelling1933analysis} based on the covariance method \cite{smith2002tutorial} . We choose to keep the first 2,000 components (out of 9,576), who's eigenvalues account for $91.6\%$ of the total variability in the data.

The training of the reservoir computer, illustrated in Fig. \ref{fig:principle}(c), was performed frame-wise on a subset of 450 video sequences, each one containing a single motion sequence; 150 sequences were used to evaluate the performance of the system. Figure \ref{fig:principle}(d) illustrates the 6 binary classifiers, introduced to distinguish the motions: 6 output nodes have been trained to give a ``1'' for each frame of the correct motion, and ``0'' for the other frames. The winner-takes-all approach, shown in Fig. \ref{fig:principle}(e), is used to classify each individual frame.
The classifier output is evaluated throughout the full video sequence (from the first frame to the last) and the final result corresponds to the class having the majority of frames within the sequence attributed to it.

During the training, the NMSE cost function (see Eq. \ref{eq:nmse}) was used to minimise the error between the reservoir output and the target class. 
In this study, it was noted that the final classification did not require the output of the correct class to be as close as possible to ``1'', while the others being close to ``0''. Since we use the winner-takes-all approach, all it takes for the correct class to ``win'' is to be slightly higher than the others. In other words, lower NMSE does not necessarily mean less classification errors. Therefore, we used a different error metric based on the confusion matrix \cite{schuldt2004recognizing}. Here, the confusion matrix is a $6\times 6$ array (dictated by the number of classes) computed for the entire video stream, each cell $p(i,j)$ giving the percentage of actions of class $i$ classified into the class $j$.
In other words, the diagonal of the confusion matrix contains the correct classification produced by the system, while non-zero elements off the diagonal correspond to errors. 
We use the confusion matrix to compute a new metric for the reservoir computer performance, called \emph{the score}, given by the sum of the diagonal elements.
A perfect classification corresponds to a score of 600, as all the 6 actions have been recognised with a 100\% accuracy.

\subsection{Photonic reservoir computer}
\label{subsec:exp}

A typical discrete-time reservoir computer contains a large number $N$ of internal variables $x_{i\in 0 \ldots N-1} (n)$ evolving in discrete time $n \in \mathbb{Z}$, as given by
\begin{equation}
  x_i(n+1) = \rev{f_{\text{NL},I}} \left( \sum_{j=0}^{N-1} W_{ij}^{res} x_j(n) + \sum_{j=0}^{K-1} B_{ij} u_j(n) \right),
  \label{eq:rcevo}
\end{equation}
where \rev{$f_{\text{NL},I}$} is a nonlinear function (in this work, \rev{$f_{\text{NL},I}(x) = \lfloor \sin^2(\lfloor x \rfloor_8) \rfloor_{10}$}), 
$W_{ij}^{res}$ is a $N\times N$ matrix of interconnecting weights between the neurons of the neural network, 
$u_j(n)$ is an input with $K$ dimensions, 
and $B_{ij}$ is a $N \times K$ matrix of input weights, often referred to as the \emph{input mask}. 
Further information on the principles of reservoir computing and the properties of $B_{ij}$ and $W_{ij}^{res}$ can be found in the Methods section.

\begin{figure}[t]
  \centering
  \includegraphics[width=0.9\textwidth]{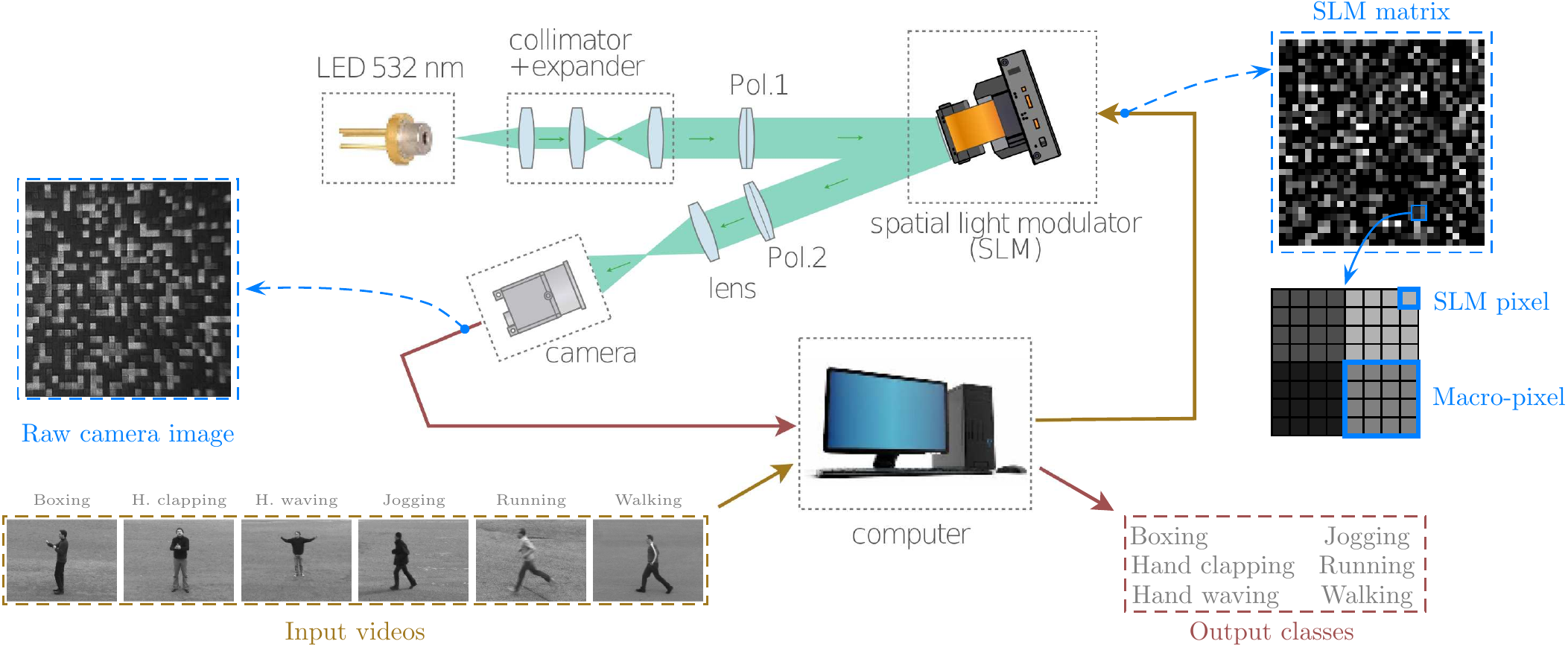}
  \caption{Illustration of the experimental setup, composed of an optical arm, connected to a computer. The output of a green LED (LED 532 nm) is collimated and expanded (collimator+expander), then polarised (Pol. 1), and used to illuminate the surface of the spatial light modulator (SLM). The latter is being imaged by a high-speed camera (camera) through a second polariser (Pol. 2) and an imaging lens (lens). Both the camera and the SLM are controlled by a computer, running a Matlab script. The latter generates the inputs from the input videos, and computes the values of pixels to be loaded on the SLM, i.e. the SLM matrix. Groups of small individual pixels of the SLM are combined into larger macro-pixels, that are easier to separate on the raw camera image. The computer uses the data from the camera to extract the reservoir states, compute the outputs and generate the output classes.}
  \label{fig:experiment}
\end{figure}

Our experimental setup implements Eq. \ref{eq:rcevo} and is schematised in Fig. \ref{fig:experiment}. It is composed of two parts: a free-space optical arm and a computer. The optical part implements the nonlinearity $f(x) = \sin^2 (x)$ in Eq. \ref{eq:rcevo}. 
It is powered by a green LED source at $532\units{nm}$ (Thorlabs M530L3) set to a power level of $10.5 \units{mW}$. 
\rev{The choice of the wavelength was based on the availability of optical components and the ease of use and calibration of the setup in the visible spectrum. The optical power was adjusted so as to provide sufficient illumination of the SLM to generate the highest contrast, with adequate exposure settings on the camera.}
The output beam is linearly polarised, collimated and expanded to roughly $17 \units{mm}$ in diameter to evenly illuminate the entire $7.68 \units{mm} \times 7.68 \units{mm}$ surface of the spatial light modulator (Meadowlark XY Phase P512 -- 0532 with 8-bit resolution). 
In simplified terms, a SLM is a variable, spatially resolved wave plate: its index of refraction along the slow axis can be decreased electronically.
That is, a linearly polarised illumination beam, parallel to the slow axis of the SLM, is reflected with a phase-only modulation.
If a beam is parallel to the fast axis instead, one would observe no modulation with the variable voltage.
In this setup, an illumination beam oriented at $45^\circ$ with respect to the slow axis provides equal optical field components to the fast and the slow axis of the SLM. After reflection, the former remains unchanged, while the latter undergoes a phase modulation.
A second polariser transforms the phase difference between the two components into intensity modulations, which are in turn imaged onto a high-speed camera (Allied Vision Mako U-130B with 10-bit resolution). The imaging-system is optimised for a compromise between imaging resolution and the field of view's extend.

The computer runs a Matlab script controlling both the SLM and the camera, taking care of loading the data into the SLM and obtaining images from the camera.
The input mask $B_{ij}$ and the interconnection matrix $W_{ij}^{res}$ are generated randomly at the beginning of the experiment. At each discrete timestep $n$, the input to the nonlinear function $\sum W_{ij}^{res} x_j(n) + \sum B_{ij}u_j(n)$ is computed, and the resulting matrix is loaded onto the SLM device. 
The camera then records a picture of the SLM through the imaging lens and the polarising optics. 
The raw image is cropped to the area of interest (the surface of the SLM) and averaged over the macro-pixels (see below), resulting in a square matrix, that represents the updated reservoir states, given by Eq. \ref{eq:rcevo}. 
The states are rearranged into a vector $x_j$ and used to compute the next SLM matrix at timestep $n+1$.

In this experiment, the reservoir size is defined by several factors. The device used here has a resolution of $512\times512$ pixels, and allows in theory for a network size of $512 \times 512 = 262,144$ neurons, if each individual pixel was used as a node. However, in our experiment this is challenging, since the SLM surface is slightly tilted with respect to the camera sensor. Consequently, only a limited region of the SLM is seen in focus by the camera, while the rest is blurry. Therefore, in this experiment, we only use the central $384 \times 384$ region of the SLM, and assign square groups of pixels, that we call macro-pixels, to individual reservoir nodes. For instance, a small network of $N=1,024$ nodes is obtained by setting the macro-pixel size to $12\times 12$, while a large network ($N=16,384$) is obtain by reducing the macro-pixel size to $3\times 3$ pixels on the SLM.

The speed of the setup is imposed by Matlab, that is, by the time needed to compute the next SLM matrix from the raw camera image.
For a large reservoir of $N=16,384$ nodes, the system is capable of processing 2 video frames per second.
In the case of a small reservoir ($N=1,024$), the matrix multiplication $\sum W_{ij}^{res} x_j (n)$ (see Eq. \ref{eq:rcevo}) requires less computations, and the processing speed is increased up to 7 frames per second.
The use of Matlab at this stage is a deliberate choice, as it lends considerable flexibility to the setup, for example testing different pre-processing techniques, reservoir topologies, and output decision-making layers by simply changing the code, i.e. without reconfiguration of the optical setup.
The system's speed limitation can be alleviated by replacing the computer with a dedicated digital signal processing (DSP) board, or a field-programmable gate array (FPGA) chip, capable of performing the matrix-products computations in real time (as in e.g. \cite{antonik2017online}).
More importantly, matrix can equally be offloaded to fully parallel optics \cite{bueno2018reinforcement, psaltis1985optical}.
\rev{As our SLM model supports refresh rates up to $300 \units{Hz}$ in overdrive mode, the hardware would be capable of processing a video stream in real time. Furthermore and because of its high frequency of operation, we could also theoretically time-multiplex up to $12$ video streams at $25$ fps.}

\subsection{Reservoir size and classification performance }
\label{subsec:size}

\begin{figure}[t]
    \centering
    \includegraphics[width=0.50\textwidth]{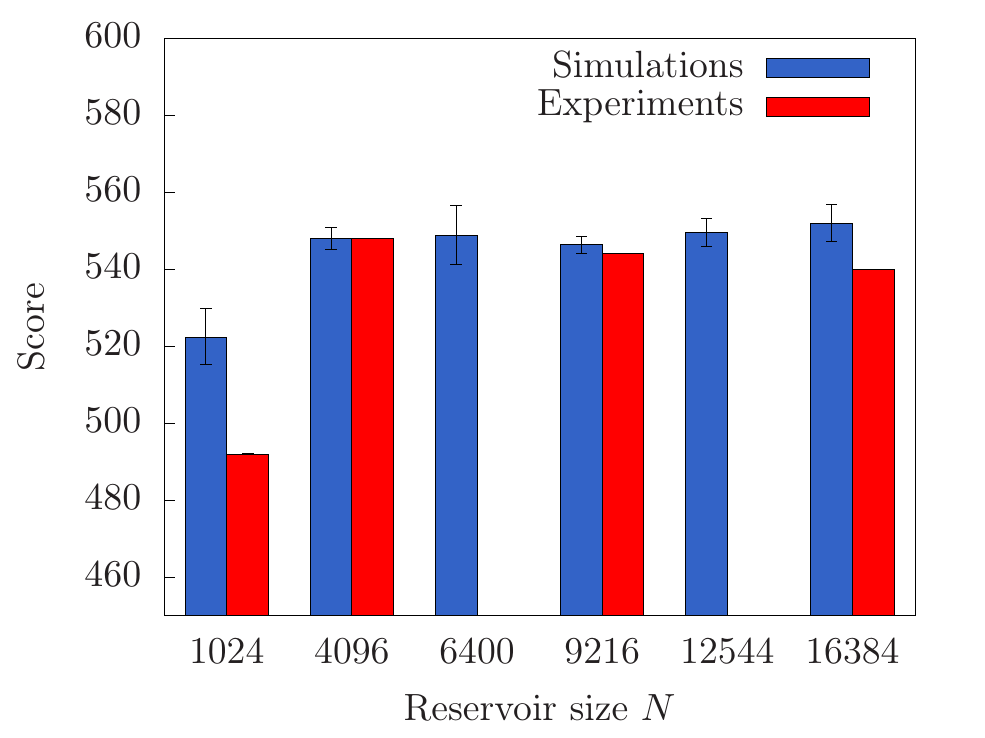}
  %
    \caption{Performance of our photonic neuro-inspired architecture on the human action classification task. Different reservoir sizes have been investigated numerically (blue) and experimentally (red). The error bars on the numerical results show the score variability (standard deviation) with 5 different input masks. Experimental variability could not be measured because of the long runtime of the experiment.}
  \label{fig:scores}
\end{figure}

To test the potential of our large-scale architecture on a challenging computer-vision task, we studied the impact of the reservoir size on the classification performance.
We investigated reservoir sizes from $N=1,024$ up to $N=16,384$, both numerically and experimentally. 
Figure \ref{fig:scores} shows the resulting performance for different reservoir sizes in terms of the classification score (introduced in Sec. \ref{subsec:kth}), computed during the testing phase.
The hyperparameters were optimised from scratch for each reservoir size, and independently in simulations and experiments. Details on the optimisation approach can be found in Sec. \ref{subsec:params}.
In numerical simulations, we investigated the performance with different random input and interconnection weights.
We performed 5 distinctive simulations for each reservoir size with different random wights and full optimisation of the hyperparameters. 
Blue bars display the average performance and the error bars show the standard deviations, i.e. the variability of the score due to different random weights.
In the experiment (red bars) such statistical analysis was hampered by the long measurement duration, from a few days for smaller reservoirs up to a week for $N=16,384$. 
\rev{Such long experimental runtimes are due to the optimisation of the hyperparameters through grid search (see Sec. \ref{subsec:params}). With one set of hyperparameters, the experiment processes the full dataset (i.e. both the training and test stages) in $1.6$ to $5.5$ hours, depending on the reservoir size.}

The graph shows a steep increase in performance from a small reservoir size of $N=1,024$ nodes up to $N=4,096$, both in numerical simulations and the experiment. The average score at $N=4,096$ is 548 in both cases. The numerical results keep improving for large reservoirs, reaching an average score of 552 at $N=16,384$. The experimental results, on the other hand, exhibit a slight decrease in performance with large reservoirs. This downturn is due to experimental imperfections, such as tilt and misalignment of the area of interest cropped from raw camera images, that become more noticeable as the macro-pixels shrink. However, the decrease has little significance, with a $1.3\%$ performance drop between $N=4,096$ and $N=16,384$.

\begin{table}[t]
  \centering
     \begin{tabular}{l|l|c|c|c|c|c}
       & & & & & \multicolumn{2}{c}{Performance} \\
       Authors & Method &  Database split & Training time & Processing speed & \emph{s1} scenario & Full database \\
      \hline
      Yadav et al. \cite{yadav2016action} & IP + SVM & 80\%-20\%& -- & -- & -- & $98.20\%$ \\
      Shi et al. \cite{shi2015learning} & DTD, DNN & 9-16 & -- & -- & -- & $95.6\%$ \\
      Kovashka et al. \cite{kovashka2010learning} & BoW + SVM & 8-8-9 & -- & -- & -- & $94.53\%$ \\
      Gilbert et al. \cite{gilbert2011action} & HCF + SVM & LOOCV & $\sim 5.6\units{h}$ & $24\units{fps}$ & -- & $94.5\%$ \\
      Baccouche et al. \cite{baccouche2011sequential} & CNN \& RNN & 16-9 & -- & -- & -- & $94.39\%$ \\
      Ali and Wang \cite{ali2014learning} & DBN \& SVM & 50\%-20\%-30\% & -- & -- & -- & $94.3\%$ \\
      Wang et al. \cite{wang2011action} & DT + SVM & 16-9 & -- & -- & -- & $94.2\%$ \\
      Liu et al. \cite{liu2008learning} & MMI + SVM & LOOCV & -- & -- & -- & $94.15\%$ \\
      Sun et al. \cite{sun2009action} & FT + SVM & auto & -- & -- & -- & $94.0\%$ \\
      Veeriah et al. \cite{veeriah2015differential} & Differential RNN & 16-9 & -- & -- & -- & $93.96\%$ \\
      Shu et al. \cite{shu2014bio} & SNN & 9-16 & -- & -- & $95.3\%$ & $92.3\%$ \\
      Laptev et al. \cite{laptev2008learning} & FT + SVM & 8-8-9 & -- & -- & -- & $91.8\%$ \\
      Jhuang \cite{jhuang2007biologically} & $StC_2$ + SVM & 16-9 & -- & $0.4\units{fps}$ & $96.0\%$ & $91.6\%$ \\
      Klaeser et al. \cite{klaeser2008spatio} & 3D Grad + SVM & 8-8-9 & -- & -- & -- & $91.4\%$ \\
      \textbf{This work} & \textbf{Photonic RC} & \textbf{75\%-25\%} & 1.6 -- 5.5 h & \textbf{$2-7\units{fps}$} & \textbf{91.3\%} & -- \\
      Grushin et al. \cite{grushin2013robust} & LSTM & 16-9 & 1 day & $12-15\units{fps}$ & -- & $90.7\%$ \\
      Ji et al. \cite{ji20133d} & 3DCNN & 8-8-9 & -- & -- & -- & $90.02\%$ \\
      Escobar et al. \cite{escobar2012action} & MT cells & 16-9 & -- & -- & $74.63\%$ & -- \\
      Schuldt et al. \cite{schuldt2004recognizing} & FT + SVM & 8-8-9 & -- & -- & -- & $71.83\%$ \\
    \end{tabular}
    \caption{\rev{Performance of various state-of-the-art digital approaches compared to our best experimental result. Database split indicates how the KTH database was split for training and testing of the system. Most studies choose to split by the number of subjects into either two groups (training and test, e.g. 16 subjects for training, 9 for the test) or three groups (training, validation and test, e.g. 8-8-9). LOOCV corresponds to leave-one-out cross validation: the system is trained on 24 subjects and tested on the remaining one. Training times and processing speeds are not discussed in most of the works, focusing on the classification performance. Some studies report specific results on the \emph{s1} scenario, considered in this work.}}
  \label{tab:perf_comp}
\end{table}

Table \ref{tab:perf_comp} compares the performance of our optical experiment with state-of-the-art digital \rev{approaches (the details can be found in the respective papers)}.
\rev{The table reports how the systems were trained on the KTH dataset (the database split), as well the training time and processing speed, wherever possible (those two metrics are very seldom reported in the literature, hence the large number of empty cells in the table).}
\rev{A few studies also report the system performance specifically on the \emph{s1} scenario, thus directly comparable to our results.}
\rev{In terms of performance, the photonic RC is short by $4.7\%$ from the best results on the \emph{s1} scenario\cite{jhuang2007biologically}, but outperforms the SVM approach in terms of processing speed by a factor of ten.}
\rev{The training time of our system is significantly shorter than deep approaches\cite{grushin2013robust}, and comparable to the SVM method with hierarchical compound features \cite{gilbert2011action}.}
\rev{Our photonic approach thus offers a performing, more flexible, and easy to train classification system. Furthermore, the recent development of integrated photonic reservoir computers\cite{vandoorne2014experimental} could give raise to very energy-efficient optical processors.}

Table \ref{tab:conf:exp:l} displays the confusion matrices for the best scores obtained numerically (552 at $N=16,384$) and experimentally (548 at $N=4,096$).
The results obtained from the experiment agree very well with the numerical simulations. In particular, we would like to point out that this does not only hold for the overall score, but also for the confusion-matrix's individual entries. This confirms the excellent controllability and robustness of the experimental system.
Specifically, hand gestures (first three rows) are perfectly recognised. 
Fast spatial movements of the subjects -- jogging and running -- are more challenging to differentiate because, for instance, one subject's running may be very similar to another subject's jogging. 
Therefore, the confusion matrices reflect several errors between these two classes. 
The walking action is also similar in appearance, but slower on the temporal scale, hence, it is more accurately classified by the system. 

\setlength\tabcolsep{0pt}
\begin{table}
  \centering
  \vspace{1.7cm}
  \begin{picture}(0,0)
    \put(68,55){\rotatebox{45}{Boxing}}
    \put(94,55){\rotatebox{45}{Hand clapping}}
    \put(120,55){\rotatebox{45}{Hand waving}}
    \put(146,55){\rotatebox{45}{Jogging}}
    \put(172,55){\rotatebox{45}{Running}}
    \put(198,55){\rotatebox{45}{Walking}}
  \end{picture}
  \begin{tabular}{l|cccccc|}
    \hhline{~------}
    Boxing & \g{100}{100} & \z & \z & \z & \z & \z \\
    Hand clapping & \z & \g{100}{100} & \z & \z & \g{0}{0} & \z \\
    Hand waving & \z & \z & \g{100}{100} & \z & \z & \z \\
    Jogging & \z & \z & \z & \g{80}{80} & \rr{8}{8} & \rr{12}{12} \\
    Running & \z & \g{0}{0} & \z & \rr{12}{12} & \g{84}{80} & \rr{4}{4} \\
    Walking & \z & \z & \z & \rr{4}{4} & \g{0}{0} & \g{96}{100} \\
    \cline{2-7}
    \multicolumn{7}{c}{ } \\
    \multicolumn{7}{c}{Simulation ($N=16,384$)}
  \end{tabular} 
  \hfill
  \begin{picture}(0,0)
    \put(68,55){\rotatebox{45}{Boxing}}
    \put(94,55){\rotatebox{45}{Hand clapping}}
    \put(120,55){\rotatebox{45}{Hand waving}}
    \put(146,55){\rotatebox{45}{Jogging}}
    \put(172,55){\rotatebox{45}{Running}}
    \put(198,55){\rotatebox{45}{Walking}}
  \end{picture}
  \begin{tabular}{l|cccccc|}
    \hhline{~------}
    Boxing & \g{100}{100} & \z & \z & \z & \z & \z \\
    Hand clapping & \z & \g{100}{100} & \z & \z & \g{0}{0} & \z \\
    Hand waving & \z & \z & \g{100}{100} & \z & \z & \z \\
    Jogging & \z & \z & \z & \g{76}{80} & \rr{8}{8} & \rr{16}{16} \\
    Running & \z & \g{0}{0} & \z & \rr{20}{20} & \g{76}{80} & \rr{4}{4} \\
    Walking & \z & \z & \z & \rr{4}{4} & \g{0}{0} & \g{96}{100} \\
    \cline{2-7}
    \multicolumn{7}{c}{ } \\
    \multicolumn{7}{c}{Experiment ($N=4,096$)}
  \end{tabular} \hspace{0.1\textwidth}
  \caption{Confusion matrices with the best performance.}
  \label{tab:conf:exp:l}
\end{table}
\setlength\tabcolsep{5pt}

\section{Discussion \& conclusion}
\label{sec:ccl}

In this work, we present a photonic video-processing system for human-action recognition. Unlike the recent advances in computer vision, relying on deep learning, we have implemented a shallow recurrent neural network -- a reservoir computer -- which not only simplifies the training process, but allows one to realise the network in hardware, such as photonic systems, inherently leveraging the parallelism of optics.
We demonstrate a highly flexible optical experiment that allows to accommodate a very large number of physical nodes ($N=16,384$), with the potential of scaling up to hundreds of thousands of nodes, thus offering considerable advantages in terms of parallelism and speed, and for realisation of the crucial vector-matrix products. \rev{The natural scalability of the proposed photonic architecture could be further exploited to process multiple video feeds in parallel by allocated various regions of the SLM screen to independent reservoir computers, each processing a specific video stream with the strategy described in this paper.}

Finally and despite the simplicity of the system, its performance on the KTH dataset is comparable to state-of-the-art deep approaches and superior to gradient-optimised LSTM networks.
Our optical information processing system is particularly well suited for the data that is already in the optical domain, such as image and video processing, studied here. This work thus proposes a hardware solution to video information processing, that could outperform deep learning in terms of training time and complexity. 

\section{Methods}

\subsection{Basic principles of reservoir computing}
\label{subsec:rc}

A typical discrete-time reservoir computer was discussed in Sec. \ref{subsec:exp}, Eq. \ref{eq:rcevo}.
The dynamics of the reservoir is determined by the matrices $W_{ij}^{res}$ and $B_{ij}$, both time-independent and drawn from a random distribution with zero mean. 
The reservoir computer produces $M$ output signals $y_i(n)$, corresponding to the $M$ output nodes (in this work, $M=6$), given by a linear combination of the states of its internal variables
\begin{equation}
  y_l(n) = \sum_{j=0}^{N-1} W_{lj}^{out} x_j (n),
  \label{eq:rcout}
\end{equation}
where $W_{lj}^{out}$ are the readout weights, trained either offline (using standard linear regression methods, such as the ridge-regression algorithm \cite{tikhonov1995numerical} used here), or online \cite{antonik2017online}, in order to minimise the normalised mean square error (NMSE) between the output signal $y(n)$ and the target signal $d(n)$, given by
\begin{equation}
  \text{NMSE} = \frac{\left\langle \left( y(n) - d(n) \right)^2 \right\rangle}{\left\langle \left( d(n) - \langle d(n) \rangle \right)^2 \right\rangle} .
  \label{eq:nmse}
\end{equation}

\subsection{Physical modeling of the photonic reservoir computer}
\label{subsec:model}
\rev{The state variable $x_i(n)$ of the $i$-th photonic neuron at discrete time step $n$ is the 10-bit quantified optical intensity $\lfloor I_i(n)\rfloor_{10}$ detected by the camera. We use the structure of the setup to determine the evolution of this state variable. It starts with a linear transformation by the network adjacency matrix and the addition of a masked input data. This relation is used  to update the 8-bit quantified phase value vector loaded in the SLM's controller according to the following equation
\begin{equation}
    \lfloor\phi_i(n+1)\rfloor_8 = \sum_{j=0}^{N-1}W_{ij}^{res}x_j(n) + \sum_{j=0}^{K-1}B_{ij}u_j(n),
\end{equation}
 with $W_{ij}^{res}$ and $B_{ij}$ the reservoir adjacency matrix and input mask, respectively.}
\rev{The phase of the $i$-th SLM's macro-pixel is nonlinearly converted into an intensity value because of the peculiar polarisation configuration of the optical arm comprising the LCoS SLM and two polarisers rotated by 45 degrees with respect to the orientation of the SLM's liquid crystals in their resting state. Using the theoretical framework of Jones calculus (see Ref.~\cite{SalehTeich19} for more details), we can easily show that $\lfloor I_i(n+1)\rfloor_{10} = \lfloor I_0\sin^2(\lfloor\phi_i(n+1)\rfloor_8)\rfloor_{10}$. Hence, the evolution equation for the $i$-th neuron's state reads 
\begin{equation}
x_{i}(n+1) = f_{\text{NL},I}\left(\sum_{j=0}^{N-1}W_{ij}^{res}x_j(n) + \sum_{j=0}^{K-1}B_{ij}u_j(n)\right),
\end{equation}
with $f_{\text{NL},I}(\cdot) = \lfloor I_0 \sin^2\left(\lfloor \cdot\rfloor_8\right)\rfloor_{10}$ the nonlinear function, $I_0$ the uniform optical intensity illuminating (and reflected from) the SLM and camera. Without loss of generality, $I_0$ can be normalised at a unitary value.
Here, a reservoir output is defined by 
\begin{equation}
    y_l(n) = \sum_{j=0}^{N-1}W_{lj}^{out}x_j(n),
\end{equation}
with $W_{out}$ the readout matrix of trainable coefficients for the 6 outputs of the reservoir (one output per action to recognise).} 

\rev{An alternative approach is to consider the 8-bit quantified, macro-pixel phase-shift $\lfloor \phi_i(n)\rfloor_8$ induced by the SLM's liquid crystals  as the state variable $x_i(n)$ of the $i$-th neuron at discrete time $n$.}
\rev{In this modelling scenario, the dynamics of the system can also read
\begin{equation}
  x_i (n+1) = \left\lfloor \sum_{j=0}^{N-1} W_{ij}^{res} f_{\text{NL},\phi} x_j (n))+ \sum_{j=0}^{K-1} B_{ij} u_j (n)\right\rfloor_8,
\end{equation}
with $f_{\text{NL},\phi}(\cdot) = \lfloor I_0 \sin^2\left( \cdot\right)\rfloor_{10}$ the nonlinear function, $I_0$ the uniform optical intensity illuminating (and reflected from) the SLM and camera. Without loss of generality, $I_0$ can be normalised at a unitary value.
In this case, the reservoir output is defined by 
\begin{equation}
    y_l(n) = \sum_{j=0}^{N-1}W_{lj}^{out}f_{NL,\phi}(x_j(n).
\end{equation}}

\subsection{Hyperparameters}
\label{subsec:params}

The dynamics of the reservoir can be optimised for a given task by tuning several control parameters. The input mask $B_{ij}$ is drawn from a random distribution over the interval $[-1,1]$ and then multiplied by a coefficient $\beta$, called the \emph{input gain}, which controls the amplitude of the external input signal, that is, the degree of perturbation of the reservoir. The generation of the interconnection matrix $W_{ij}^{res}$ requires two additional parameters: a \emph{scaling factor} $\gamma$ and a \emph{density} $\rho$. Since the echo-state network paradigm \cite{jaeger2001echo} requires the interconnection matrix to be sparse, $W_{ij}^{res}$ is generated from a random distribution over the interval $[-1,1]$ with $\rho \times N^2$ non-zero elements. The matrix is then multiplied by a global scaling factor $\gamma$, which determines the strength of connections between different neurons within the network. The diagonal elements of $W_{ij}^{res}$, which define the feedback of each neuron to itself, are defined separately. Since we want all neurons to exhibit the same internal dynamics, we set the diagonal elements of $W_{ij}^{res}$ to $\alpha$, a parameter called the \emph{feedback gain}. 

In summary, the dynamics of the system are defined by four hyperparameters -- the input gain $\beta$, the feedback gain $\alpha$, the interconnection gain $\gamma$, and the interconnection density $\rho$. The optimisation of hyperparameters is performed through grid search (\textit{i.e.} parameter sweep) -- an exhaustive search through all possible combinations of manually specified values of all the parameters.
Table \ref{tab:hyperparams} presents the intervals used for the optimisation, and the optimal values for selected reservoir sizes, considered both numerically and experimentally. 

Hyperparameters optimisation have shown the input and feedback gains to be important variables, i.e. accurate values are required to obtain the best performance, while the characteristics of interconnection matrix play a minor role. We managed to obtain comparable scores with significantly different $W_{ij}^{res}$ matrices in terms of density and amplitude of the off-diagonal elements.

\begin{table}
  \centering
  \begin{tabular}{|l|c|c|c|c|c|c|}
    \hline
    \multirow{3}{*}{Parameter} & \multirow{3}{*}{Symbol} & \multirow{3}{*}{Search values} & \multicolumn{2}{c|}{Optimal for} & \multicolumn{2}{c|}{Optimal for} \\
    & & & \multicolumn{2}{c|}{$N=1,024$} & \multicolumn{2}{c| }{$N=16,384$} \\
    \cline{4-7}
    & & & Num & Exp & Num & Exp \\
    \hline
    Feedback gain & $\alpha$ & $0.1 - 1.5$ & $0.8$ & $0.8$ & $0.6$ & $0.3$ \\
    Input gain & $\beta$ & $0.0001 - 1$ & $0.01$ & $0.1$ & $0.16$ & $0.16$ \\
    Interconnectivity gain & $\gamma$ & $0.0001 - 1$ & $0.1$ & $0.1$ & $0.001$ & $0.001$ \\
    Interconnectivity density & $\rho$ & $0.0001 - 1$ & $0.01$ & $0.001$ & $0.001$ &$0.001$ \\
    \hline
  \end{tabular}
  \caption{Optimal hyperparameters for reservoirs of different sizes.}
  \label{tab:hyperparams}
\end{table}

\subsection{The KTH dataset}
\label{subsec:kth_data}

\begin{figure}[t]
  \centering
  \begin{subfigure}[t]{0.45\textwidth}
    \includegraphics[width=0.30\textwidth]{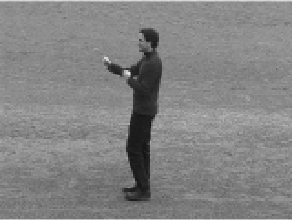}
    \hfill
    \includegraphics[width=0.30\textwidth]{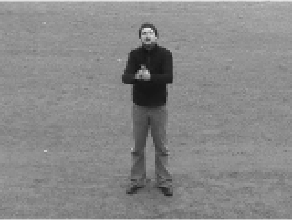}
    \hfill
    \includegraphics[width=0.30\textwidth]{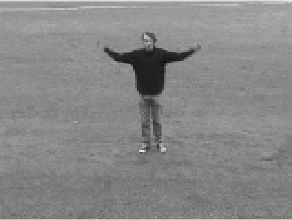}
    \includegraphics[width=0.30\textwidth]{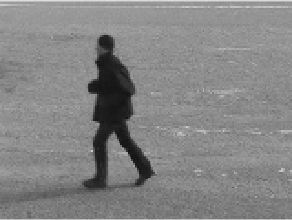}
    \hfill
    \includegraphics[width=0.30\textwidth]{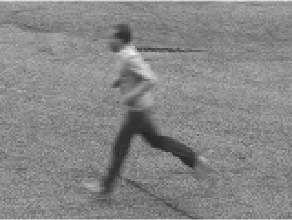}
    \hfill
    \includegraphics[width=0.30\textwidth]{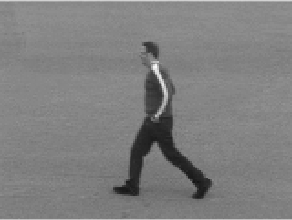}
    \caption{Examples of KTH frames}
    \label{fig:kth}
  \end{subfigure}
  \begin{subfigure}[t]{0.45\textwidth}
    \centering
    \raisebox{-1.8cm}{
      \includegraphics[width=0.60\textwidth]{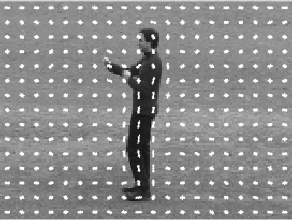}
    }
    \caption{Example of HOG features}
    \label{fig:hog}
  \end{subfigure}
  \caption{\textbf{(a)} Examples of action frames from the KTH database, from left to right: boxing, hand clapping, hand waving, jogging, running, and walking. Six different subjects are illustrated out of the total of 25. All videos have been taken outdoors over a homogeneous background, which corresponds to the ``s1'' subset of the full database.
    \textbf{(b)} Example of HOG features computed in Matlab for a frame of the KTH dataset. 
    The HOG features are visualised using a grid of rose plots. 
    The grid dimensions ($20\times 15$ here) are determined by the ratio between the image and cell sizes.
    Each rose plot shows the distribution of gradient orientations within a HOG cell. 
    The length of each petal of the rose is proportional to the contribution of each orientation within the histogram.
    The plot thus displays the edge directions, which are normal to the gradient directions.
    In this example, it allows to capture the pose of the subject.
    }
\end{figure}

The original KTH video database \cite{schuldt2004recognizing} contains four different scenarios. In this work, for simplicity, we limited the dataset to the first scenario, referred to as ``s1'', containing outdoor videos (illustrated in Fig. \ref{fig:kth}). 
All videos were recorded over homogeneous background with a static camera and $25 \units{fps}$, then downsampled to the spatial resolution of $160\times120$ pixels. Each single action movie has a length of four seconds in average. 
The subjects repeat each action 4 times. 
In total, our dataset contains $25 \times 6 \times 4 = 600$ sequences for each combination of 25 subjects, 6 actions, and 4 repetitions. 
The DIVX-compressed videos are first uncompressed and split into $160 \times 120$ grayscale frames.
Different sequences vary in length and contain between $24$ and $239$ frames.

\subsection{Histograms of oriented gradients}
\label{subsec:hog}

The histograms of oriented gradients (HOG) algorithm, introduced by Dalal and Triggs \cite{dalal2005histograms}, is based on Scale-Invariant Features transform (SIFT) descriptors \cite{lowe2004distinctive}. 
To calculate a HOG descriptor, first, horizontal and vertical gradients are computed by filtering the image with the following kernels \cite{bahi2015robust}:
\begin{equation}
  G_x = \left( -1, 0, 1 \right) \hspace{2cm} \text{and} \hspace{2cm} G_y = \left( \begin{array}{c} -1 \\ 0 \\ 1 \end{array} \right).
  \label{eq:kernels}
\end{equation}
Then, magnitude $m(x,y)$ and orientation $\theta(x,y)$ of gradients are computed for each pixel, using
\begin{equation}
  m(x,y) = \sqrt{ D_x^2 + D_y^2} \hspace{2cm} \text{and} \hspace{2cm}
  \theta(x,y) = \arctan \left( \frac{D_y}{D_x} \right),
\end{equation}
where $D_x$ and $D_y$ are the approximations of horizontal and vertical gradients, respectively.

The creation of histograms starts with the division of the image into small cells. Each cell is assigned a histogram of typically 9 bins, corresponding to angles $0, 20, 40, \ldots 160$, and containing the sums of magnitudes of the gradients within the cell.
The main purpose of this operation is to provide a compact, yet truthful description of a patch of an image. That is, a typical cell of $8\times 8$ grayscale pixels is described with 9 numbers instead of 64. As gradients of an image are sensitive to overall lighting, the algorithm is completed with block normalisation, by dividing the histograms by their euclidean norm computed over bigger-sized blocks. 

The computation of HOG features was performed in Matlab, using the built-in \pf{extractHOGFeatures} function, individually for each frame of every sequence, with a cell size of $8\times 8$ and a block size of $2\times 2$. Given the frame size of $160 \times 120$ pixels, the function returns $19 \times 14 \times 4 \times 9 = 9576$ features per frame. Figure \ref{fig:hog} illustrates the resulting gradients superimposed on top of a video-frame from the KTH dataset.

\section*{Data availability statement}

The KTH dataset can be downloaded here: 
\url{http://www.nada.kth.se/cvap/actions/}.
\rev{The numerical and experimental data can be downloaded here:
\url{add link}.}

\section*{Code availability statement}
\rev{The code used in this study can be downloaded here: \url{add link}.}




\bibliography{refs.bib}

%


\section*{Acknowledgements} 

The authors thank 
the creators of the KTH dataset for making the videos publicly available.
This work was supported by AFOSR (grants No. FA-9550-15-1-0279 and FA-9550-17-1-0072), R\'egion Grand-Est, and the Volkswagen Foundation via the NeuroQNet Project.

\section{Author contributions statement}

\rev{D.B, N.M., and D.R designed and managed the study. P.A., N.M., and D.R. realised the experimental setup. P.A. performed the numerical simulations and the experimental campaigns. P.A., N.M, and D.R. prepared the manuscript. All authors discussed the results and reviewed the manuscript. }

\section{Additional information}


\textbf{Competing interests}
The authors declare no competing interests.




\end{document}